\documentclass{article}

\usepackage{PRIMEarxiv}

\usepackage[utf8]{inputenc} 
\usepackage[T1]{fontenc}    
\usepackage{hyperref}       
\usepackage{url}            
\usepackage{booktabs}       
\usepackage{amsfonts}       
\usepackage{nicefrac}       
\usepackage{microtype}      
\usepackage{lipsum}
\usepackage{fancyhdr}       
\usepackage{graphicx}       

\usepackage{underscore}
\usepackage{algorithm}
\usepackage[noend]{algpseudocode}

\usepackage{array}

\floatname{algorithm}{Algorithm}

\graphicspath{{media/}}     

\title{RTHDet: Rotate Table Area and Head Detection in images
}

\author{
  Wenxing Hu \\
  School of Electronics \& Information Engineering\\
  Shanghai University of Electric Power\\
  Shanghai 200090, China \\
  \texttt{moonstarwork@gmail.com} \\
   \And
  Minglei Tong \\
  School of Electronics \& Information Engineering\\
  Shanghai University of Electric Power\\
  Shanghai 200090, China \\
  \texttt{tongminglei@gmail.com} \\
}

\begin{document}
\maketitle

\begin{abstract}

Traditional models and datasets primarily concentrated on horizontal table detection, unable to accurately detect table regions and localize their head and tail parts in rotating scenarios, thus greatly restricting the development of table recognition in rotated contexts. Therefore, this paper presents a new task of detecting table regions and localizing their head and tail parts in rotation scenarios, along with the proposal of corresponding datasets, evaluation metrics, and methods. An innovative method——Adaptively Bounded Rotation, is introduced to address the lack of datasets for detecting rotated table and their head-tail parts. Through this method, this study has produced TRR360D, a rotated table detection dataset which incorporates semantic information of table head and tail, based on the classic horizontal table detection dataset, ICDAR2019MTD. Furthermore, the study proposes a new evaluation metric, R360 AP, to validate the precision of rotated region detection and head-tail localization, building on the classic evaluation metric, AP. Following extensive literature review and rigorous experimentation, the high-speed and high-accuracy rotated detection model, RTMDet-S, was selected as the baseline. Furthermore, this study proposed the RTHDet model. RTHDet defines the $D_{r360}$ rotated rectangle angle representation based on the baseline and applies it to the newly added AL (Angle Loss) branch. This enhancement allows the model to locate the head and tail of the rotated table. By applying transfer learning methods and adaptive boundary random rotation data augmentation techniques, the AP50 (T<90) of RTHDet has been improved from 23.7\% to 88.7\% compared to the baseline model. This validates the effectiveness of RTHDet in handling the novel task of detecting rotating table regions and accurately localizing their head and tail parts.RTHDet is integrated into the widely-used open-source MMRotate toolkit: \url{https://github.com/open-mmlab/mmrotate/tree/dev-1.x/projects/RR360}.

\end{abstract}


\section{Introduction}

The vast amount of data from the objective world is concealed in ubiquitous paper documents and forms, such as financial statements, express waybills, medical examination reports, and exam papers. Extracting valuable data from these tables requires the development of theories and technologies for table digitization and intelligence. This is crucial for enhancing efficiency in finance, education, and logistics industries, and for accelerating their intellectual development. However, when the paper world is converted into digital images, various rotate, affine, perspective transformations, and geometric distortions emerge, posing significant challenges for the localization and analysis of table regions.

Traditional models and datasets mainly focus on the detection of horizontal tables as shown in Fig.\ref{dtask} (a), failing to precisely detect table regions in rotated scenarios as depicted in Fig.\ref{dtask} (b), and even more incapable of locating the head and tail of tables as illustrated in Fig.\ref{dtask} (c). This greatly restricts the development of downstream tasks in table recognition OCR. Therefore, this paper introduces a new task for detecting table regions in rotated scenes and locating heads and tails, and conducts in-depth research on it. The main contributions of this paper in proposing the new task of rotated table region detection and head-tail localization are as follows:

\begin{figure}[ht]

\centering
\includegraphics[width=16.5cm]{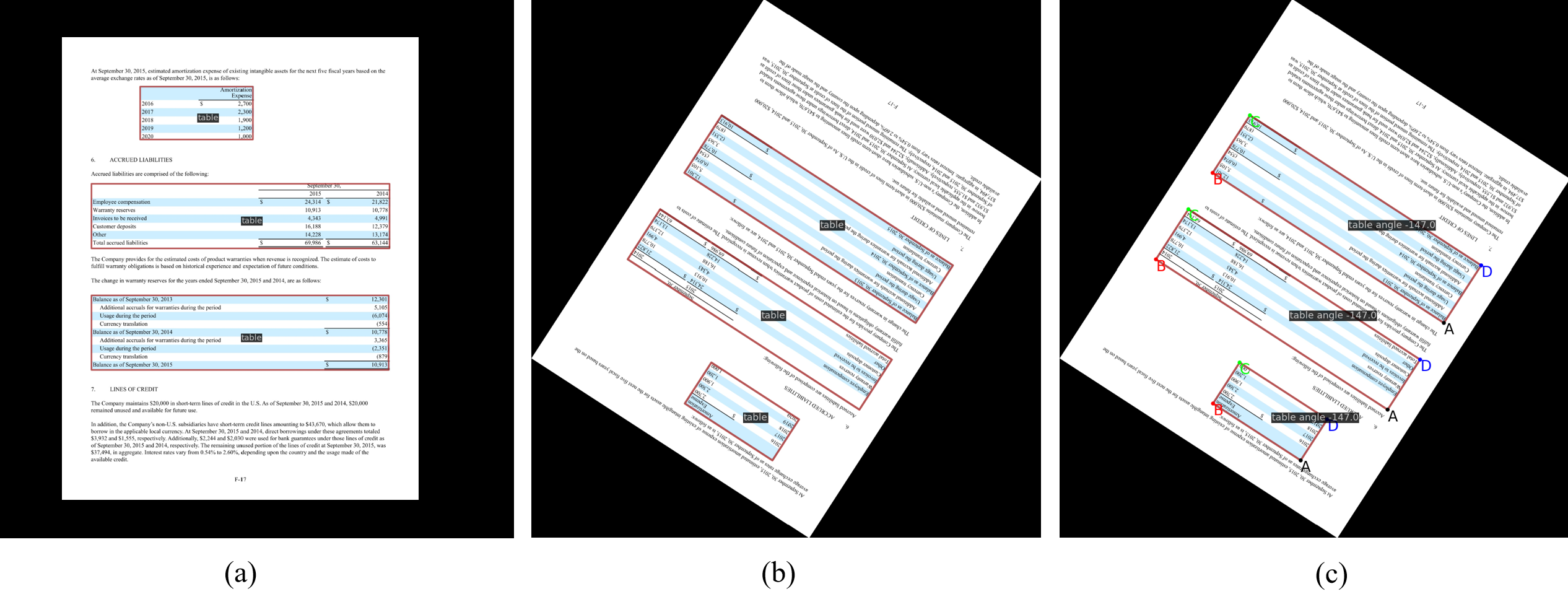}

\caption{Table Detection Task: (\textbf{a}) Detection of Table Horizontal Regions (\textbf{b}) Detection of Table Rotation Regions (\textbf{c}) Detection of Table Head and Tail \label{dtask}}
\end{figure}

(1) TRR360D Dataset: In response to the lack of a dedicated dataset for rotated table images in existing table detection research, this paper presents a novel dataset generation method called Adaptively Bounded Rotation. Drawing upon the ICDAR2019MTD\cite{Gao2019} modern table detection dataset, the QBox (Quadrangle Box) annotation format from the DOTA\cite{Xia2018} dataset is adopted to represent rotated boxes. By applying the Adaptively Bounded Rotation technique, the starting point and annotation direction of the QBox are regulated, thereby generating the TRR360D rotating table detection dataset, which inherently carries semantic information of the table's head and tail. This dataset comprises 840 images with 1446 rotated box annotations and is publicly available at \url{https://github.com/vansinhu/TRR360D}.

(2) R360 AP Evaluation Metric: To address the limitation of the traditional AP evaluation metric, which only measures the accuracy of rotating table region predictions but does not account for the precision of head and tail localization, this paper introduces the R360 AP evaluation metric. Building upon the AP metric, the R360 AP metric incorporates angle constraints to assess the performance of both rotating table region detection and head-tail localization accuracy.

(3) RTHDet Model: In response to the issue of existing models being unable to distinguish between the head and tail of rotated tables, the RTHDet (Rotate Table Head Detector) model is introduced. This model builds upon the RTMDet-S \cite{Lyu2022} baseline model and proposes the $D_{r360}$ rotation rectangle box angle representation method. This method is applied to the novel AL (Angle Loss) branch, enabling the detection of table regions and head-tail parts in rotating scenarios. By utilizing transfer learning and adaptive boundary random rotation data enhancement techniques, the RTHDet model achieves an AP50(T<90) score of 88.7\%, representing a substantial improvement of 65.0 percentage points compared to the RTMDet-S baseline model. This demonstrates RTHDet's ability to effectively detect rotating table regions and precisely localize their head and tail parts. RTHDet is integrated into the widely-used open-source MMRotate toolkit: \url{https://github.com/open-mmlab/mmrotate/tree/dev-1.x/projects/RR360}.

The structure of the paper is as follows: the first section serves as an introduction, the second section covers the related work, the third section presents the methodology, the fourth section discusses the experimental results, and the fifth section provides a summary of the entire paper.

\section{Related Work}

In this section, an overview of the related work on horizontal table detection in document images is provided, categorized into stages of rule-based methods and deep learning-based methods. The limitations of horizontal table detection algorithms in accurately detecting rotated table regions and localizing their head and tail are analyzed. Furthermore, a brief introduction is given to the impact of classical and state-of-the-art methods for detecting rotated regions and their limitations when directly applied to table detection in the field.

\subsection{Horizontal Table Detection}

\subsubsection{Rule-Based Method}

In early-stage research, rule-based methods were commonly employed. Itonori et al. \cite{itonori1993table} addressed the issue of table detection in document images using a rule-based method. Their approach harnessed the arrangement of text blocks and the position of ruling lines to identify tables within documents. Chandran and Kasturi \cite{chandran1993structural} proposed an alternative method focusing on ruling lines for table detection. Similarly, Pyreddy and Croft \cite{pyreddy1997tintin} developed a heuristics-based table detection method that initially identified structural elements from a document and subsequently filtered the tables.

Researchers have also developed tabular layouts and grammar for detecting tables in documents \cite{green1995recognition}. One such method predicts tables using the correlation of white spaces and vertical connected component analysis \cite{hu1999medium}. Another approach, introduced by Pivk et al. \cite{pivk2007transforming}, converts tables found in HTML documents into a logical structure. Shigarov et al. \cite{shigarov2016configurable} exploit metadata from PDF files, treating each word as a text block. Their method restructures tabular boundaries by leveraging bounding boxes for each word. For a more in-depth understanding of these rule-based methods, readers are encouraged to consult references \cite{embley2006table}.  

While early rule-based systems can detect tables in documents with limited patterns, they rely on manual intervention to search for optimal rules. Furthermore, their propensity to produce generic solutions poses a challenge.Rule-based methods are not sufficiently robust and are limited to specific scenarios, unable to handle the challenges posed by diverse table styles. In contrast, deep learning-based approaches, as described in the following sections, effectively overcome these limitations in horizontal table detection.

\subsubsection{Deep Learning Method}

With the advancement of deep learning, an increasing number of researchers have turned to it to address horizontal table detection challenges.The progress of object detection networks in computer vision has been shown to have a direct correlation with advancements in table detection in document images. Gilani et al. \cite{Gilani2017} demonstrated the applicability of Faster R-CNN \cite{Ren} to table detection in document images, treating it as an object detection problem. Their work involved using distance transform methods to modify the pixels in raw document images fed to the Faster R-CNN.

Building upon this foundation, Schreiber et al. \cite{schreiber2017deepdesrt} proposed another method that utilized Faster R-CNN \cite{Ren}, in conjunction with pre-trained base networks, such as ZFNet \cite{schreiber2017deepdesrt} and VGG-16 \cite{Simonyan2015}, to detect tables in document images. Siddiqui et al. \cite{Siddiqui2018} also developed a Faster R-CNN-based approach that implemented deformable convolutions \cite{Dai2017} to tackle table detection with arbitrary layouts. In addition, Reference \cite{Sun2019} adopted a Faster R-CNN with a corner-correlating approach to enhance the prediction of tabular boundaries in document images.

Saha et al. \cite{saha2019graphical} conducted an empirical study that revealed Mask R-CNN \cite{He2020} outperforms Faster R-CNN in detecting tables, figures, and formulas. This finding was further corroborated by Zhong et al. \cite{zhong2020image} who used Mask R-CNN to localize tables. Additionally, YOLO \cite{redmon2018yolov3}, SSD \cite{Liu2016}, and RetinaNet \cite{lin2017focal} have been employed to demonstrate the advantages of closed-domain fine-tuning for table detection in document images.

More recently, cutting-edge object detection algorithms, such as Cascade Mask R-CNN \cite{cai2018cascade}, Hybrid Task Cascade (HTC) \cite{chen2019hybrid} and DetectoRS\cite{Qiao2020a}, have been incorporated to enhance the performance of table detection systems in document images \cite{Prasad2020,Agarwal2020,Zheng2021,Nazir2021,CasTabDetectoRS}. Despite the advancements made by these prior methods, there remains significant potential for improving the localization of accurate tabular boundaries in scanned document images. Furthermore, existing table detection techniques often rely on more substantial backbones or memory-intensive deformable convolutions. Ma et al. enhanced Faster R-CNN's accuracy for table detection by leveraging CornerNet in their RobusTabNet \cite{MA2023109006} method, demonstrating superior performance on various benchmarks with a basic ResNet-18 network.

These methods primarily support the detection of horizontally aligned tables but do not cater to rotated table detection. They also lack the capability to differentiate between the head and tail sections of tables.

\subsection{Rotate Table Area Detection}\label{subsection:relateadefine}

Traditional object detection algorithms, such as RetinaNet\cite{lin2017focal} and CascadeRCNN\cite{Cai2021}, can only detect the horizontal bounding boxes of table objects, and cannot detect the rotated bounding boxes of tables. In recent years, the field of rotated object detection has also made significant progress. Subsequently, an introduction to the representation of horizontal objects and the representation of rotated object regions will be provided. This will be followed by an analysis of the limitations of existing rotation detection algorithms when applied to table region detection.

\begin{figure}[ht]
\centering
\includegraphics[width=1.0\textwidth]{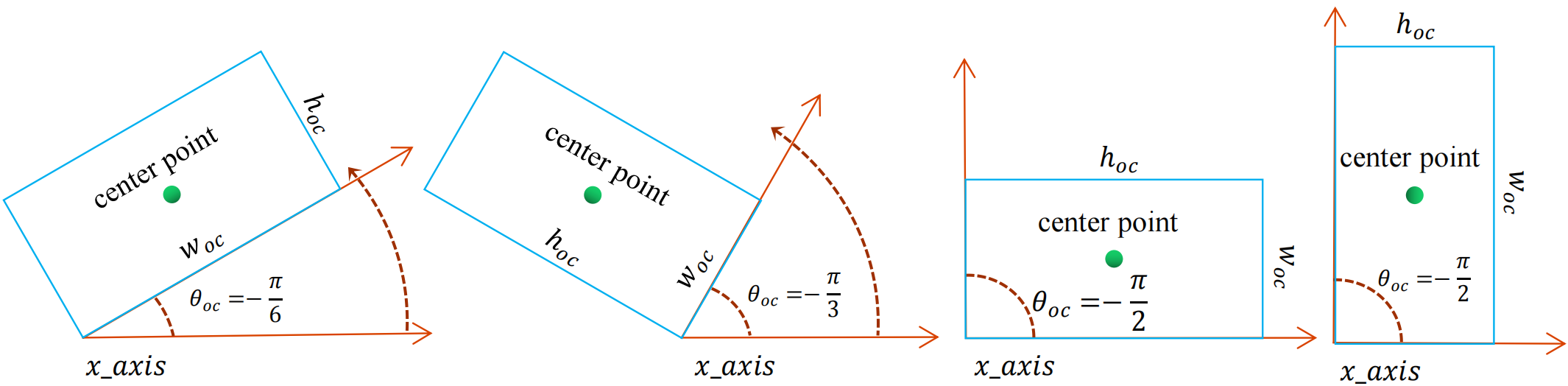}
\caption{$D_{oc}$:the old version OpenCV definition,when $OpenCV<4.5.1$, $angle \in [-90,0 ^\circ)$, $\theta \in [-\frac{\pi}{2},0)$, and the angle between the width and the positive x-axis is a positive acute angle. At this time, the width edge will exchange as the angle changes. This definition comes from the cv2.minAreaRect function in OpenCV, which returns a value in the range $[-90^\circ, 0)$.\label{oc1}}
\end{figure}

\begin{figure}[ht]
\centering
\includegraphics[width=1.0\textwidth]{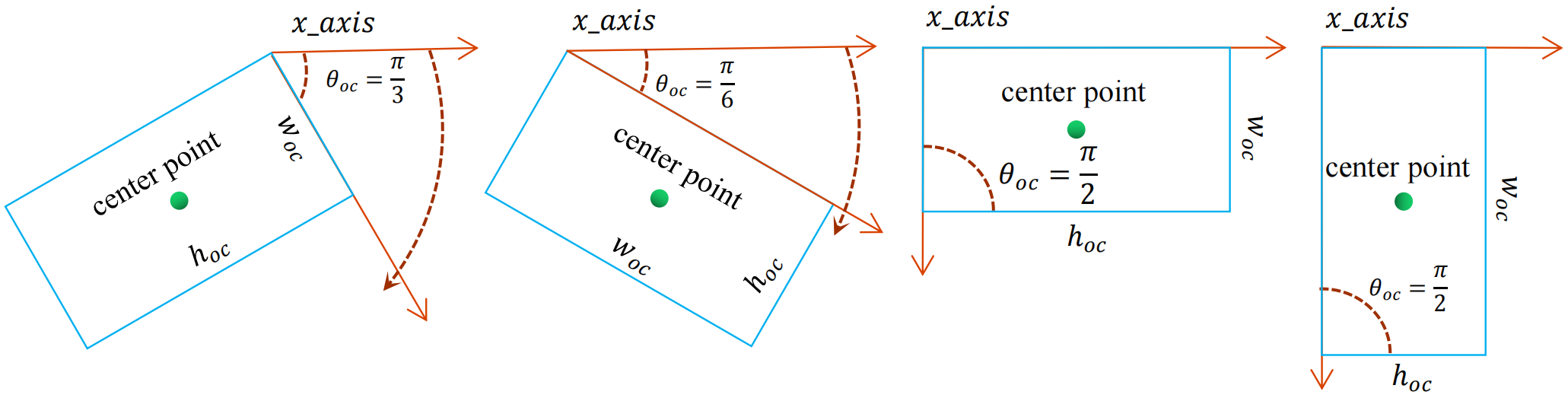}
\caption{$D_{oc}'$:the new version of OpenCV,when $OpenCV\geq4.5.1$, $angle \in(0,90^\circ]$,$\theta \in (0,\frac{\pi}{2}]$, and the angle between width and the positive x-axis is acute (positive), based on the $cv2.minAreaRect$ function in OpenCV, which returns a value in the range of $(0,90^\circ]$.\label{oc2}}
\end{figure}

\begin{figure}[ht]
\centering
\includegraphics[width=1.0\textwidth]{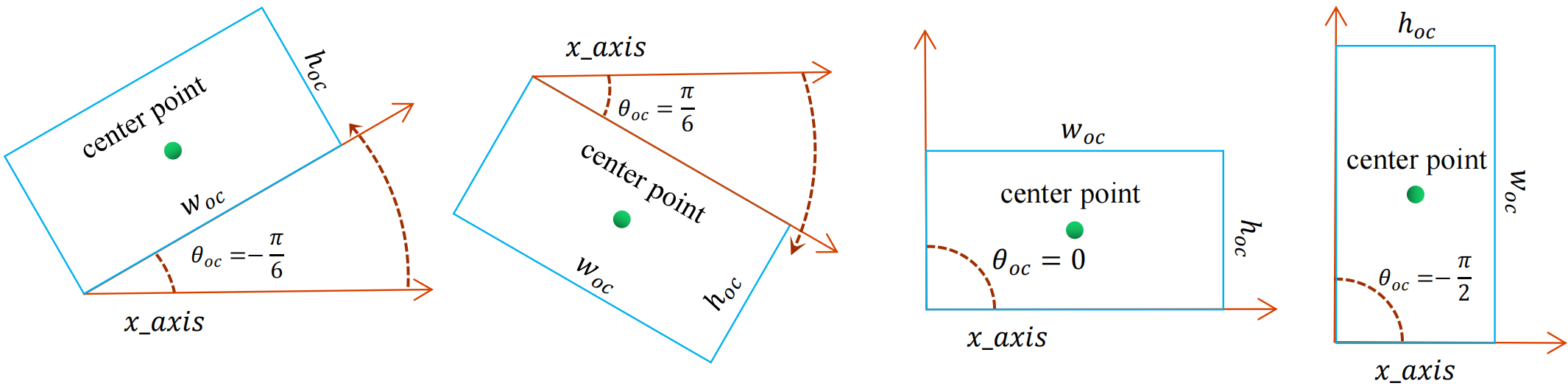}
\caption{$D_{le90}$: Long Edge 90 Definition, $angle\in[-90^\circ,90^\circ]$, $\theta\in[-\frac{\pi}{2},\frac{\pi}{2}]$, and $width>height$.\label{le90}}
\end{figure}

\begin{figure}[ht]
\centering
\includegraphics[width=1.0\textwidth]{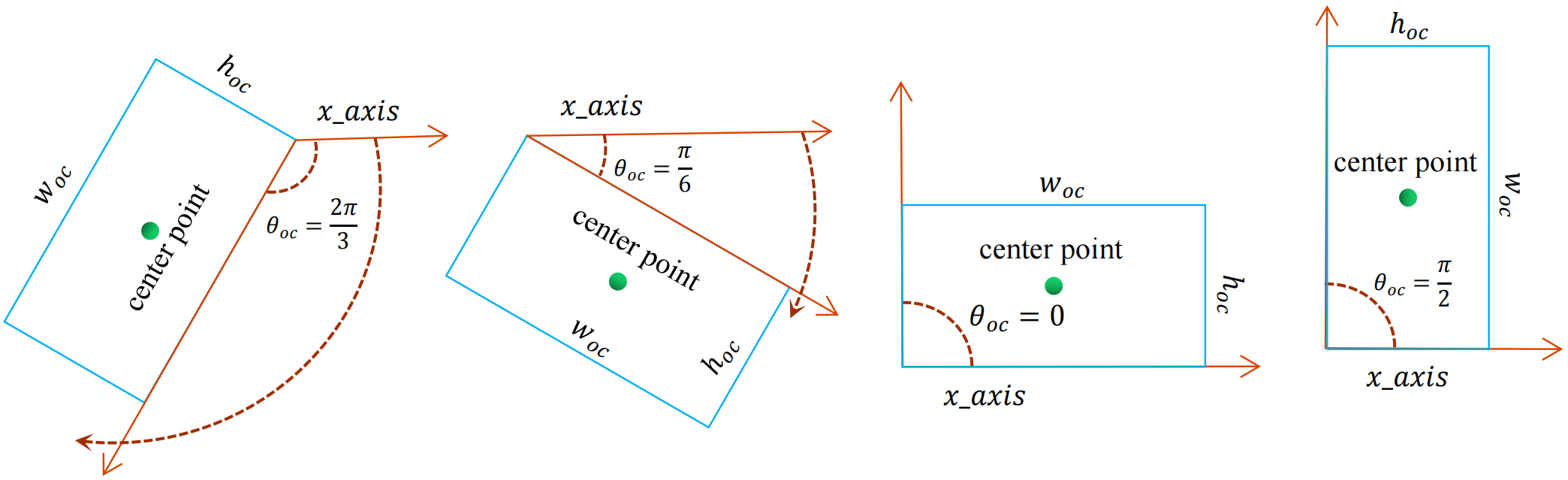}
\caption{$D_{le135}$: Long Edge 135 Definition,$angle \in [-45^\circ,135^\circ]$, $\theta \in [-\frac{\pi}{4},\frac{3\pi}{4}]$, and $width>height$.\label{le135}}
\end{figure}

Traditional object detection algorithms predict parameters using center width and height representation, denoted as $[x, y, w, h]$. However, in the field of rotated table detection, an additional parameter $\theta$ is required. Previous work on the DOTA dataset used four ways to define $\theta$: $D_{oc}$ shown in Figure \ref{oc1}, $D_{oc}'$ shown in Figure \ref{oc2}, $D_{le90}$ shown in Figure \ref{le90}, and $D_{le135}$ shown in Figure \ref{le135}. Although these representations can represent the rotated region of the table, they cannot convey the semantics of the four corners of the rotated region, nor can they represent the head and tail of the table.

Therefore, the rotation region detection models in MMRotate, such as RetinaNet-R, CascadeRCNN-R, and RTMDet-S, suffer from the limitation of angle definition, which prevents them from accurately localizing the head and tail parts of the rotated regions. However, the semantic information of the head and tail parts is crucial for improving the accuracy of downstream tasks like table recognition and OCR.In Section \ref{subsection:r360}, a new angle representation will be proposed and applied to existing rotation region detection models to enable the localization of the head and tail parts. Thus, it propels the development of downstream tasks.


\section{Method}

By conducting experiments as described in Section \ref{subsection:ebaseline}, the RTMDet-S model was selected as the baseline for rotating table region detection due to its superior speed and accuracy.The RTMDet model is an improved version of the object detection algorithm YOLOX\cite{Ge2021}, which is very similar to YOLOX in overall structure and consists of three parts: CSPNeXt, CSPNeXtPAFPN, and SepBNHead. CSPNeXt and CSPNeXtPAFPN are used to extract features, and SepBNHead is used for detection. Like YOLOX, the core module in RTMDet is also CSPLayer, but the Basic Block is improved to CSPNeXt Block.

\begin{figure}[ht]
\centering
\includegraphics[width=1.0\textwidth]{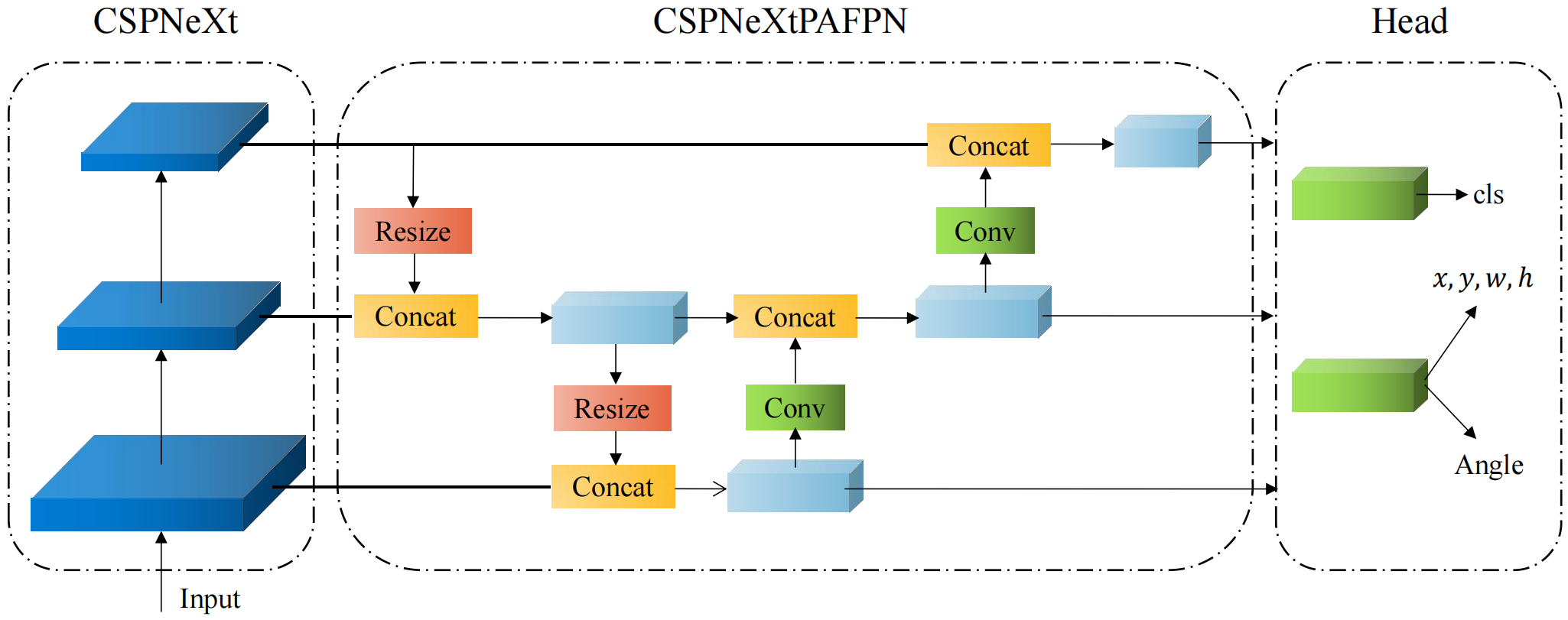}
\caption{RTHDet Diagram\label{RTMDet}}
\end{figure}

To address the limitation of the baseline model in localizing the head and tail parts of rotated regions, a new angle definition method is proposed in this section, which is applied to the AL (Angle Loss) branch of the RTMDet-S model. The resulting model is named RTHDet, as illustrated in Figure \ref{RTMDet}. Additionally, the Adaptively Bounded Rotation method is proposed for the generation of the dataset in Section \ref{subsection:trr360d} and for random data augmentation during the training process.

\subsection{Definition of R360 Rotation Box Angle}\label{subsection:r360}

\begin{figure}[H]
\centering
\includegraphics[width=10.5 cm]{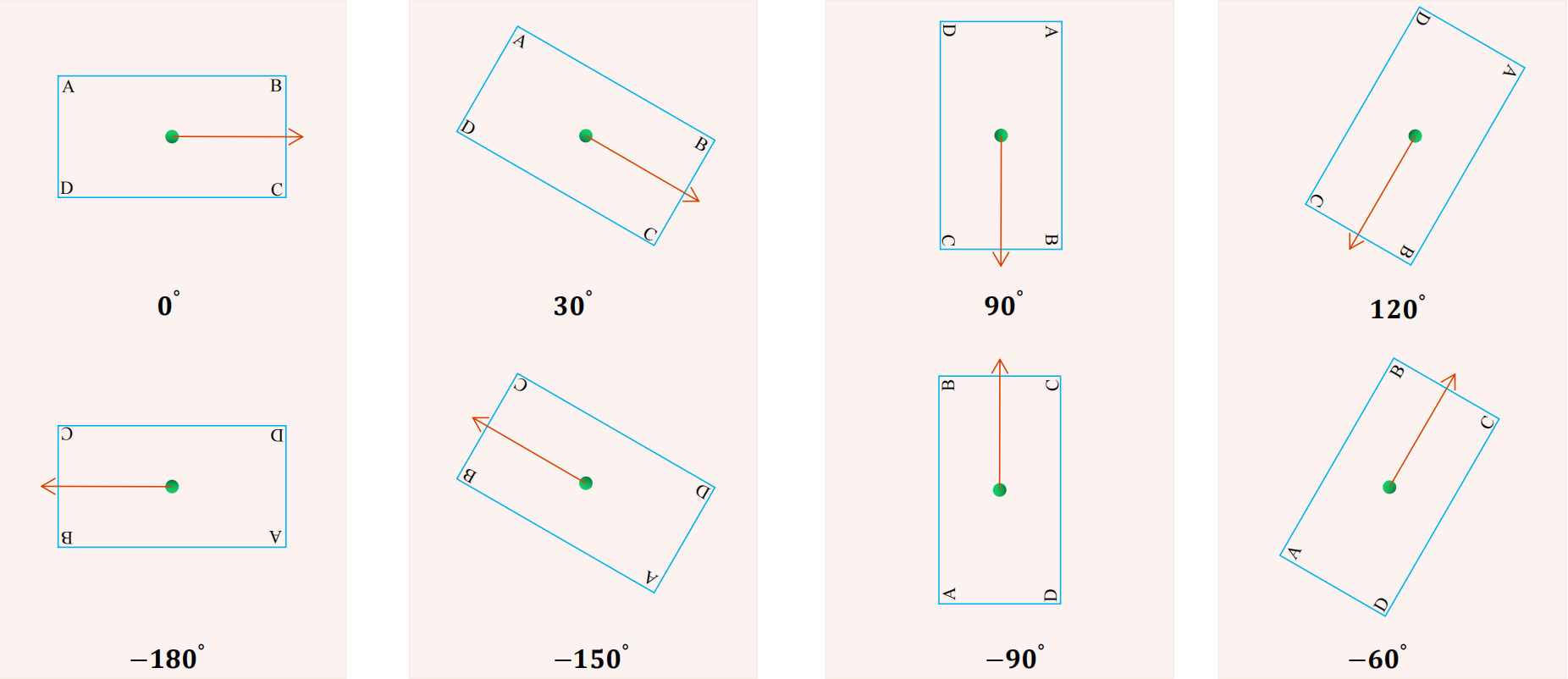}
\caption{ $D_{r360}$: $angle \in [-180^\circ,180^\circ)$,$\theta \in [-\pi,\pi)$ \label{fig:r360}}
\end{figure}

The four angle definition methods mentioned in section \ref{subsection:relateadefine}, namely $D_{oc}$, $D_{oc}'$, $D_{le90}$, and $D_{le135}$, are unable to distinguish between the head and tail of objects, such as tables. Additionally, the angle range that can be represented is limited to $180^\circ$, meaning that it cannot distinguish between 0° and -180°, as illustrated in Figure \ref{fig:r360}. To address these limitations, this study introduces the RBox (Rotated Bounding Box) using the R360 angle definition method, denoted as $D_{r360}$. The Rbox of a rotated rectangle bounding box is defined by Equation \ref{equa:rboxd}, with an angle range of [-180°, 180°). Furthermore, the corresponding QBox quadrilateral bounding box for RBox can be defined using Equation \ref{equa:qboxd}.

\begin{equation}
	RBox=(cx,cy,w,h,\theta)
 \label{equa:rboxd}
\end{equation}

\begin{equation}
	QBox=[(x_A,y_A),(x_B,y_B),(x_C,y_C),(x_D,y_D)]
 \label{equa:qboxd}
\end{equation}

\subsubsection{QBox2RBox}

As shown in Equation \ref{equa:minAreaRect}, the $cv2.minAreaRect$ function in OpenCV is able to calculate the center point $(cx, cy)$ and the width and height $(w, h)$ of a rotated rectangle, given a set of points. However, the resulting angle range is limited to $angle \in (0, 90^\circ]$, which corresponds to the angle definition in the new version of OpenCV denoted as $D_{oc'}$. This method is only capable of representing the rotated area of a table and cannot be used to distinguish between the table head and tail.

\begin{equation}
	((cx,cy),(w,h),angle)=cv2.minAreaRect(QBox)
 \label{equa:minAreaRect}
\end{equation}

Therefore, in this research, the value of $\theta$ is calculated using the inverse trigonometric functions in numpy as shown in Equation \ref{equa:npartan2}. The domain of $\theta$ is $[-\pi, \pi)$, expressed in radians.

\begin{equation}
	\theta=np.arctan2(y_C-y_B,x_C-x_B)
 \label{equa:npartan2}
\end{equation}

\subsubsection{RBox2QBox}

\begin{figure}[ht]
\centering
\includegraphics[width=11.5 cm]{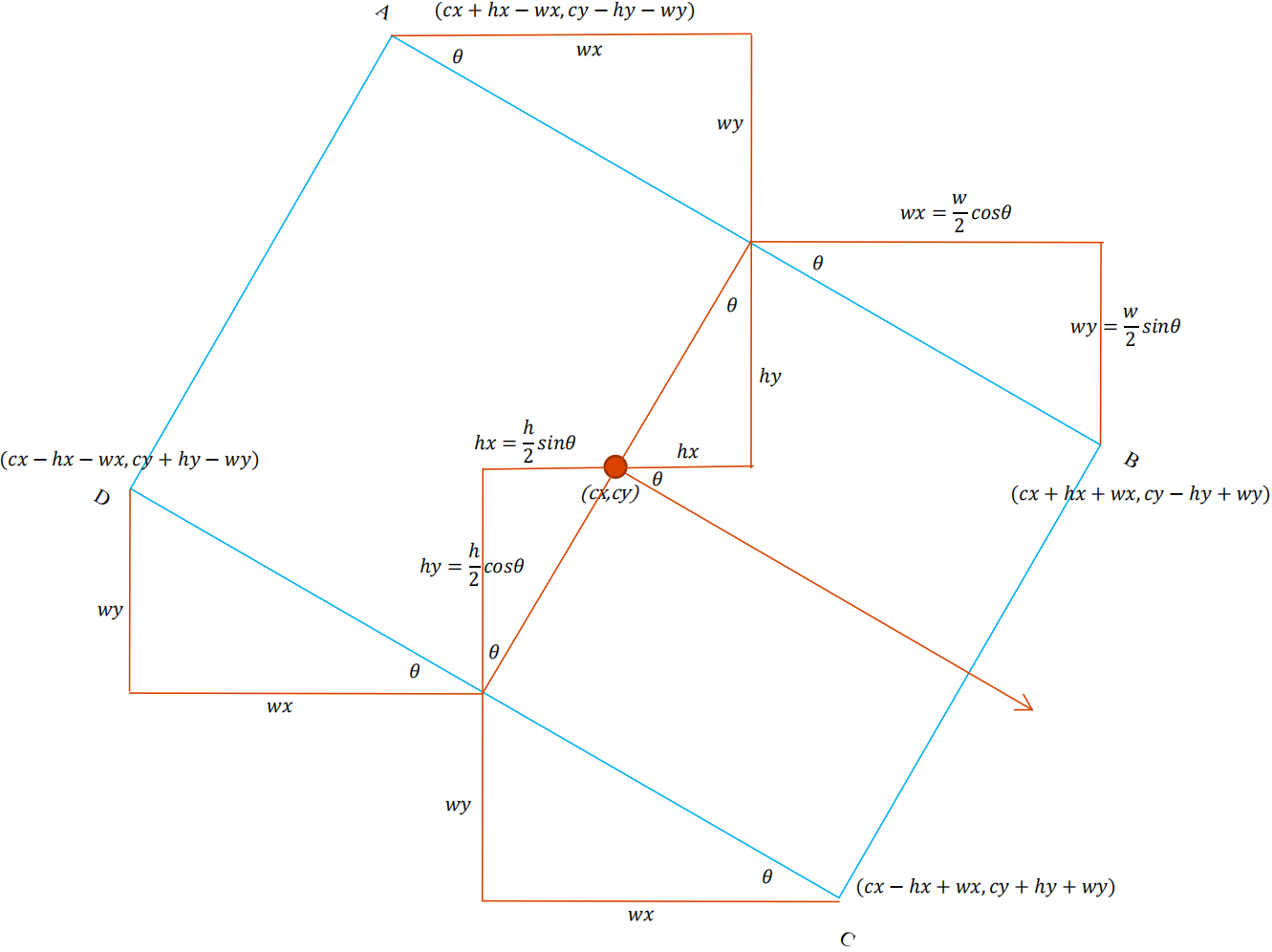}
\caption{Relationship between RBox and Qbox \label{fig:RQBox}}
\end{figure}

The relationship between RBox and QBox is shown in Figure \ref{fig:RQBox}, and all the parameters of QBox can be obtained from RBox through Equation \ref{equa:QBoxA}, \ref{equa:QBoxB}, \ref{equa:QBoxC}, and \ref{equa:QBoxD}.

\begin{equation}
(x_{A}, y_{A})=(cx+\frac{h}{2}sin\theta-\frac{w}{2}cos\theta,cy-\frac{h}{2}cos\theta-\frac{w}{2}sin\theta)
 \label{equa:QBoxA}
\end{equation}

\begin{equation}
(x_{B}, y_{B})=(cx+\frac{h}{2}sin\theta+\frac{w}{2}cos\theta,cy-\frac{h}{2}cos\theta+\frac{w}{2}sin\theta)
 \label{equa:QBoxB}
\end{equation}

\begin{equation}
(x_{C}, y_{C})=(cx-\frac{h}{2}sin\theta+\frac{w}{2}cos\theta,cy+\frac{h}{2}cos\theta+\frac{w}{2}sin\theta)
 \label{equa:QBoxC}
\end{equation}

\begin{equation}
(x_{D}, y_{D})=(cx-\frac{h}{2}sin\theta-\frac{w}{2}cos\theta,cy+\frac{h}{2}cos\theta-\frac{w}{2}sin\theta)
 \label{equa:QBoxD}
\end{equation}

\subsection{AL Branch}

The RTMDet algorithm has four models, namely L (Large), M (Medium), S (Small), and T (Tiny). The baseline model of RTMDet in MMRotate uses the rotated IoU loss function shown in Equation \ref{equa:lossiou} to regress the rotated bounding boxes (RBox), where RIoU is illustrated in Figure \ref{fig:rotatediou}. However, the limitation of $\mathcal{L}_{Rbox}$ is that it can only learn the features of rotated regions and cannot learn the $360^\circ$ rotation angle features that include table head and tail information.

\begin{equation}
	\mathcal{L}_{RBox} = 1-RIoU
 \label{equa:lossiou}
\end{equation}

To facilitate the learning of 360° rotation features, an Angle L1 Loss (AL) branch was incorporated. In conjunction with the $D_{r360}$ rotation rectangle box angle definition method defined earlier, the AL branch employs the loss function depicted in Equation \ref{equa:lossangle} to constrain angle learning. This adaptation enables the RTMDet model to support $360^\circ$ rotation table detection and consequently head-tail location.

\begin{equation}
	\mathcal{L}_{Angle} = \left| Angle_{P} - Angle_{G} \right|
 \label{equa:lossangle}
\end{equation}

\subsection{Adaptively Bounded Rotation}\label{subsection:AdaptivelyBoundedRotation}

\begin{figure}[ht]
    \centering
    \includegraphics[width=0.9\textwidth]{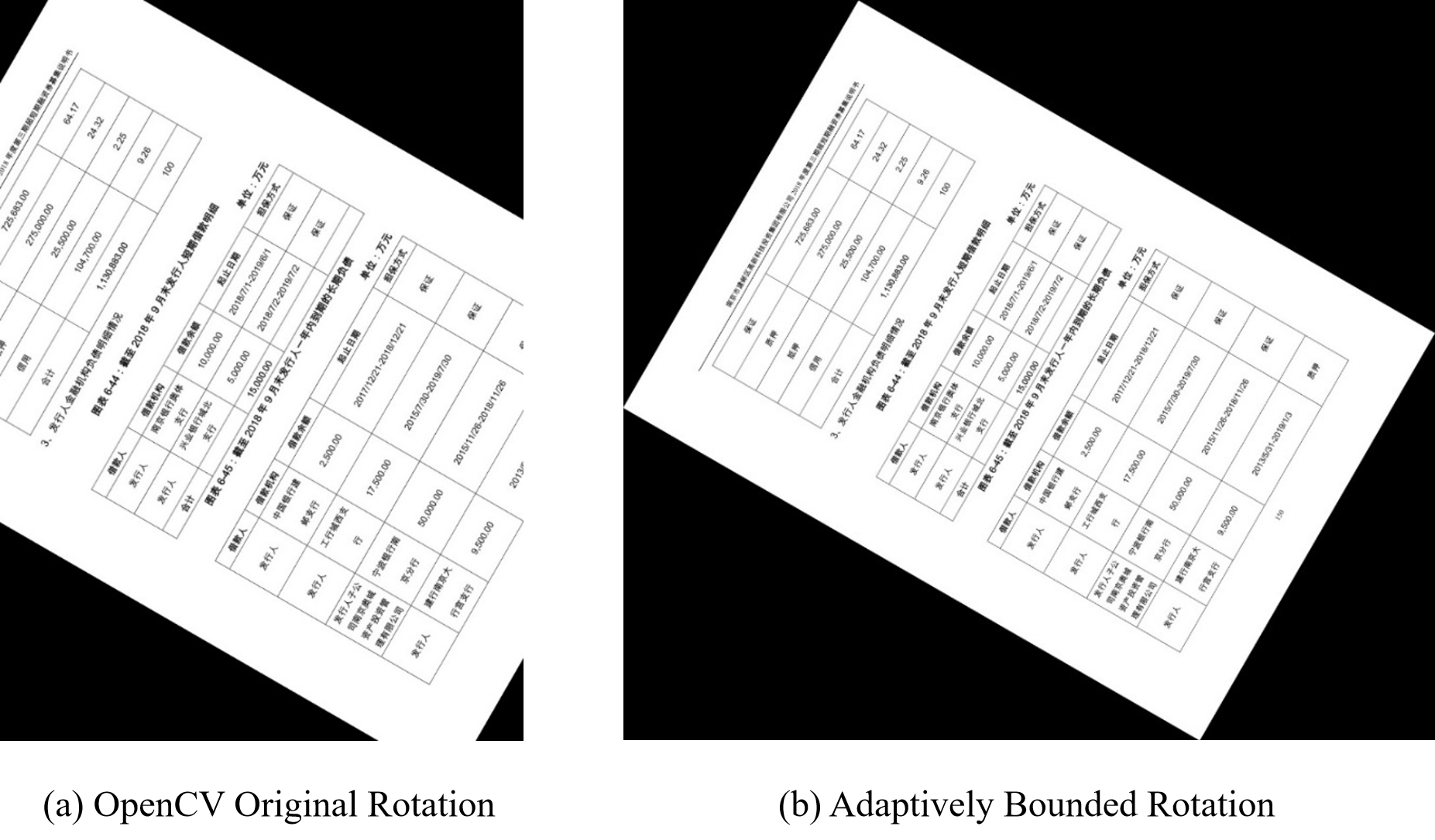}
    \caption{Rotation transformation}
    \label{fig:Rotate}
\end{figure}

The original OpenCV rotation algorithm \ref{algorithm:a1} suffers from the issue of table image region loss, as illustrated in Figure \ref{fig:Rotate} (a). To address this issue, an adaptive boundary rotation transformation algorithm is proposed in this study. The algorithm flow is illustrated in Algorithm \ref{algorithm:a2}, and the transformation effect is displayed in Figure \ref{fig:Rotate} (b).

\begin{algorithm}
\caption{OpenCV Original Rotation}
\begin{algorithmic}[1]
\Require image, $\theta$, points: List[(x,y)]
\Ensure r\_image, r\_points: List[(x,y)]

\State h$\gets$image.height
\State w$\gets$image.weight
\State matrix$\gets$cv2.getRotationMatrix2D(center, -angle, 1)

\State r_image$\gets$cv2.warpAffine(image, matrix, (w, h))
\State pts $\gets$ points.reshape([-1, 2])
\State pts $\gets$ np.hstack([pts, np.ones([len(pts), 1])]).T
\State points $\gets$ matrix@points
\State r_points $\gets$ [[points[0][x],points[1][x]] for x in range(len(points[0]))]

\State \textbf{return} r\_image, r\_points
\end{algorithmic}
\label{algorithm:a1}
\end{algorithm}

\begin{algorithm}[ht]
\caption{Adaptively Bounded Rotation}
\begin{algorithmic}[1]
\Require image, $\theta$, points: List[(x,y)]
\Ensure r\_image, r\_points: List[(x,y)]

\State h$\gets$image.height
\State w$\gets$image.weight
\State matrix$\gets$cv2.getRotationMatrix2D(center, -angle, 1)

\State cos = abs(matrix[0, 0]) 
\State sin = abs(matrix[0, 1])
\State new_w = h * sin + w * cos
\State new_h = h * cos + w * sin
\State matrix[0, 2]$\gets$matrix[0, 2]+(new_w - w) * 0.5
\State matrix[1, 2]$\gets$matrix[1, 2]+ (new_h - h) * 0.5
\State r_image$\gets$cv2.warpAffine(image, matrix, (new\_w, new\_h))

\State pts $\gets$ points.reshape([-1, 2])
\State pts $\gets$ np.hstack([pts, np.ones([len(pts), 1])]).T
\State points $\gets$ matrix@points

\State r_points $\gets$ [[points[0][x],points[1][x]] for x in range(len(points[0]))]

\State \textbf{return} r\_image, r\_points
\end{algorithmic}
\label{algorithm:a2}
\end{algorithm}

\section{Experiment}

This section is divided into three major parts. Firstly, it introduces a new dataset and evaluation metrics specifically designed for the task of detecting rotated table regions and localizing their head and tail parts. Secondly, it presents the experiments conducted to select a baseline model for rotated table region detection. Lastly, it showcases the ablation experiments conducted to investigate the localization of table head and tail parts.

\subsection{Dataset}

In response to the scarcity and high annotation cost associated with rotated image table detection datasets, particularly for head/tail detection, this study introduces a method for constructing a comprehensive rotated image table detection dataset. 

\subsubsection{ICDAR2019MTD} 

\begin{figure}[ht]
    \centering
    \includegraphics[width=16.5cm]{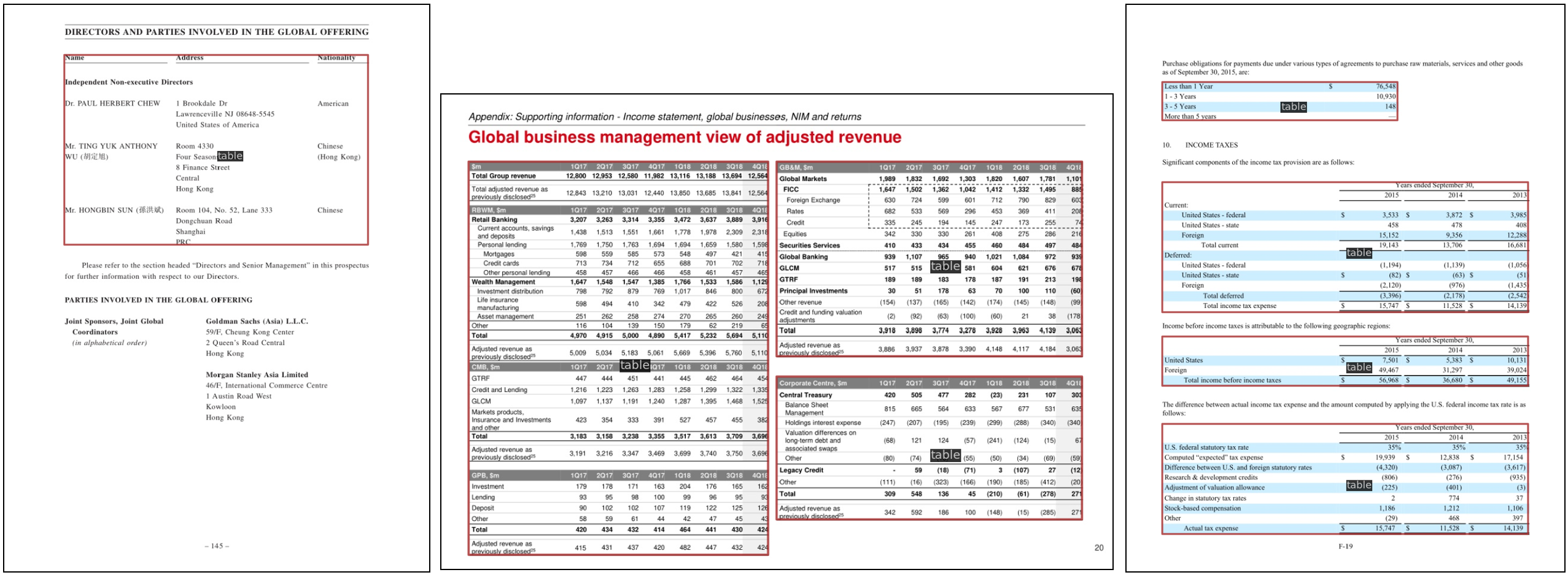}
    \caption{ICDAR2019MTD Visualization}
    \label{fig:orinann}
\end{figure}

The ICDAR2019MTD\cite{Gao2019} Modern Table Detection dataset was introduced at the Table Detection and Recognition Competition of the 2019 International Conference on Document Analysis and Recognition. Comprising 600 training images and 240 testing images, it contains 977 annotated table instances in the training set and 449 annotated table instances in the testing set, provided in XML format. The annotation format for this dataset adheres to the convention displayed in Equation \ref{equa:icdar2019}, with the top-left corner of the image serving as the starting point and the remaining annotation points arranged counterclockwise. Figure \ref{fig:orinann} illustrates the ICDAR2019MTD visualization without any head and tail information.

\begin{equation}
	x_{1} \ y_{1} \ x_{2} \ y_{2} \ x_{3} \ y_{3} \ x_{4} \ y_{4}
\label{equa:icdar2019}
\end{equation}

A notable limitation of the ICDAR2019MTD dataset is its exclusive focus on horizontally-aligned tables, rendering it unsuitable for training table rotation object detectors. Moreover, the four-point annotation format lacks semantic information and does not specify the starting point as the top-left corner of the table object. To capitalize on the capabilities of the MMRotate\cite{Zhou2022} rotation object detection algorithm, this study converts the original ICDAR2019MTD dataset annotations in XML format to DOTA-format\cite{Xia2018} text annotations, while introducing constraints and methods for generating TRR360D annotations.

\subsubsection{TRR360D} \label{subsection:trr360d}

\begin{figure}[ht]

    \centering
    \includegraphics[width=16.5cm]{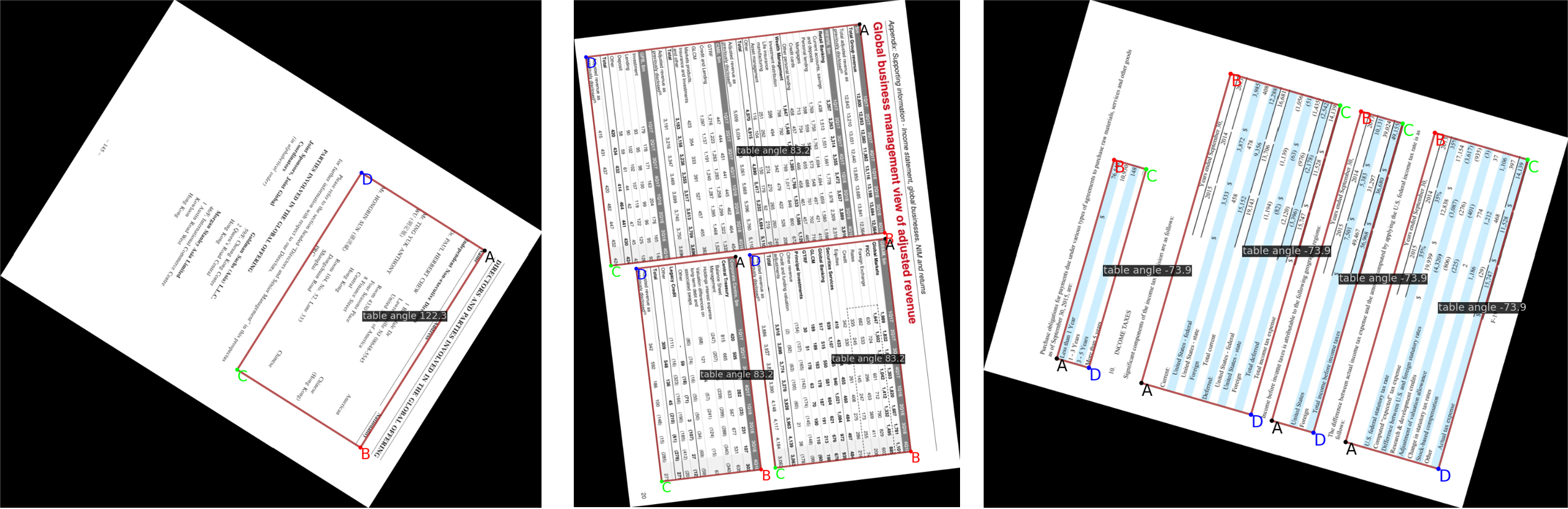}
    \caption{TRR360D Visualization}
    \label{fig:Rotate1}
\end{figure}

For the new task of detecting rotated table regions and localizing their head and tail parts, a manual adjustment was performed on a small portion of annotations in the ICDAR2019MTD dataset. Subsequently, the Adaptively Bounded Rotation algorithm \ref{algorithm:a2} was utilized to generate a rotated table detection dataset, TRR360D, which contains semantic information about the table head and tail parts.

The following provides an introduction to the format of the TRR360D dataset and the minimal manual adjustments required for annotation. Following the DOTA dataset annotation format, a line in the text file corresponding to the annotation of a rotated table instance is illustrated in Equation \ref{equa:TRR360D}. Point A represents the top-left corner of the table, and points ABCD are arranged clockwise. Parameter D denotes the detection difficulty of the sample, which is uniformly defined as 0 in TRR360D.

\begin{equation}
	x_{A} \ y_{A} \ x_{B} \ y_{B} \ x_{C} \ y_{C} \ x_{D} \ y_{D} \ table \ D 
 \label{equa:TRR360D}
\end{equation}

\begin{figure}[ht]

    \centering
    \includegraphics[width=16.5cm]{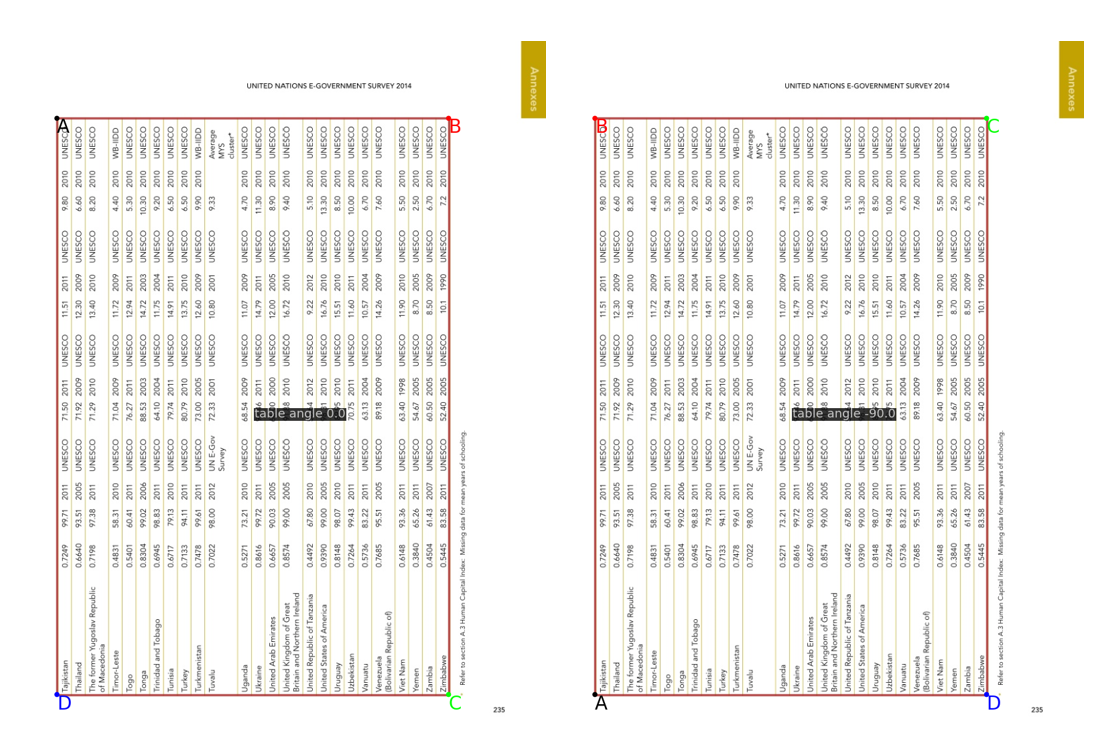}

    \caption{Adjustment of starting point labeling}
    \label{fig:AdjustPoint}
\end{figure}

The left image in Figure \ref{fig:AdjustPoint} depicts the visualization of the original annotation for image 10497 in the ICDAR2019MTD modern table detection dataset, where the four-point coordinates of ABCD are displayed as Equation \ref{equa:beforeadjust1}. Since point A is not located in the top-left corner of the table, the labeling points need to be adjusted such that point A is positioned in the top-left corner of the table, and points BCD fulfill the clockwise constraint.

\begin{equation}
	63 \ 119 \ 666 \ 119 \ 666 \ 1006 \ 63 \ 1006 \ table \ 0
 \label{equa:beforeadjust1}
\end{equation}

By manually adjusting the coordinates according to the order of Equation \ref{equa:afteradjust}, the resulting visualization is depicted in the right image of Figure \ref{fig:AdjustPoint}. Now, point A is positioned in the top-left corner of the table, and points BCD meet the clockwise constraint.

\begin{equation}
	63 \ 1006 \ 63 \ 119 \ 666 \ 119 \ 666 \ 1006 \ table \ 0
 \label{equa:afteradjust}
\end{equation}

In the original dataset, some images do not meet the constraints. Consequently, the starting point positions of images 10001, 10037, 10149, 10206, and 10211 in the testing set were manually adjusted. Furthermore, the point order of images 10062, 10108, 10187, 10418, 10445, 10497, and 10537 in the training set were manually modified to comply with the constraints of point A being located at the top-left corner of the table and points ABCD arranged in a clockwise manner.

\begin{table}
	\caption{TRR360D dataset folders and annotations}
	\centering
	\begin{tabular}{lllll}
		\cmidrule(r){1-5}
		Folder     & Description     & Format & Images & Instances\\
		\midrule
        ann\_test\_hbbox & Horizontal test set annotations & txt & 240 & 449 \\
        ann\_test\_obbox & Rotated test set annotations & txt & 240 & 449 \\
        ann\_train\_hbb & Horizontal training set annotations & txt & 600 & 977 \\
        ann\_train\_obbox & Rotated training set annotations & txt & 600 & 977 \\
		\bottomrule
	\end{tabular}
	\label{tab:ann}
\end{table}


\subsubsection{Evaluation}

To address the limitations of the traditional evaluation metric AP, which only measures the accuracy of rotation table region prediction without considering the precision of header and footer localization, this paper introduces the R360 AP evaluation metric. R360 AP is derived from AP by incorporating angle constraints and is denoted as AP50(T<90), among others.

First, the concepts of $Rotate \ IoU$, $Precision$, and $Recall$ on which the definition of AP50(T<90) relies will be introduced. Finally, a detailed description of AP50(T<90) will be provided.

\begin{figure}
    \centering
    \includegraphics[width=0.3\textwidth]{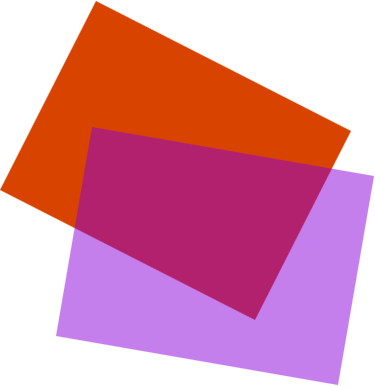}
    \caption{Rotate IoU}
    \label{fig:rotatediou}
\end{figure}

$Rotate \ IoU:$ Let the bounding boxes detected by a deep learning model be denoted as predicted boxes $P$, and the annotated boxes in the dataset as ground truth boxes $G$. The formula for Rotated IoU is given by Equation \ref{equa:riou}, as illustrated in Figure \ref{fig:rotatediou}.

\begin{equation}
	IoU_{PG}=\frac{P \cap G}{P \cup G}
 \label{equa:riou}
\end{equation}

The definitions of the subsequent metrics, such as $TP$, $Precision$, $Recall$, $F1Score$, and $AP$, are all associated with the $IoU$ between the predicted bounding boxes $P$ and the ground truth boxes $G$.

$TP$: $True \ Positive$ refers to the predicted boxes $P$ that satisfy the conditions $IoU_{PG}>T_{IoU}$ and $\left|P_{\theta}-G_{\theta}\right|<T_{\theta}$. In this case, $T_{IoU}$ represents the IoU threshold, which is only counted once for each ground truth box. A larger threshold implies a higher challenge for localization accuracy. For instance, in PASCAL VOC 2007, $T_{IoU}=0.5$. Additionally, $T_{\theta}$ denotes the angle threshold, with smaller angles corresponding to greater difficulty.

$FP$: $False \ Positive$ refers to the number of predicted boxes $P$ that satisfy $IoU \leq T_{IoU}$ or $|\theta_P - \theta_G| \geq T_{\theta}$, or the number of redundant predicted boxes detected for the same ground truth box. This metric is also known as the count of false detections.

$FN$: $False \ Negative$, the number of ground truth boxes G that were not detected by the model, also known as missed samples.

$Precision$: The ratio of the number of correctly predicted boxes to the total number of predicted boxes.

\begin{equation}
	Precision=\frac{TP}{TP+FP}
 \label{equa:Precision}
\end{equation}

$Recall$: The ratio of correct predictions to the total number of ground truth boxes is the evaluation metric for the dataset.

\begin{equation}
	Recall=\frac{TP}{TP+FN}
 \label{equa:Recall}
\end{equation}


$PR \ Curve: $The PR curve is a common performance evaluation metric used to measure the performance of object detection models at different recall and precision levels. The PR curve is plotted by calculating the recall and precision of the object detection model at different confidence threshold levels. The AP value is the area under the PR curve, which is usually computed using the 11-point method for faster computation in practical implementation.

$AP50(T<90)$ refers to the area under the precision-recall (PR) curve at a specific configuration, where the true positive (TP) condition is defined as $IoU_{PG}>0.5$ and $\left|P_{\theta}-G_{\theta}\right|<90$, where $T_{IoU}=0.5$ and $T_{\theta}=90$.

$AP75(T<40)$ refers to the area under the precision-recall (PR) curve at a specific configuration, where the true positive (TP) condition is defined as $IoU_{PG}>0.75$ and $\left|P_{\theta}-G_{\theta}\right|<40$, where $T_{IoU}=0.75$ and $T_{\theta}=40$.

\subsection{Experiment on Rotate Table Area Detection}\label{subsection:ebaseline}

\begin{figure}[H]
\centering
\includegraphics[width=16.5cm]{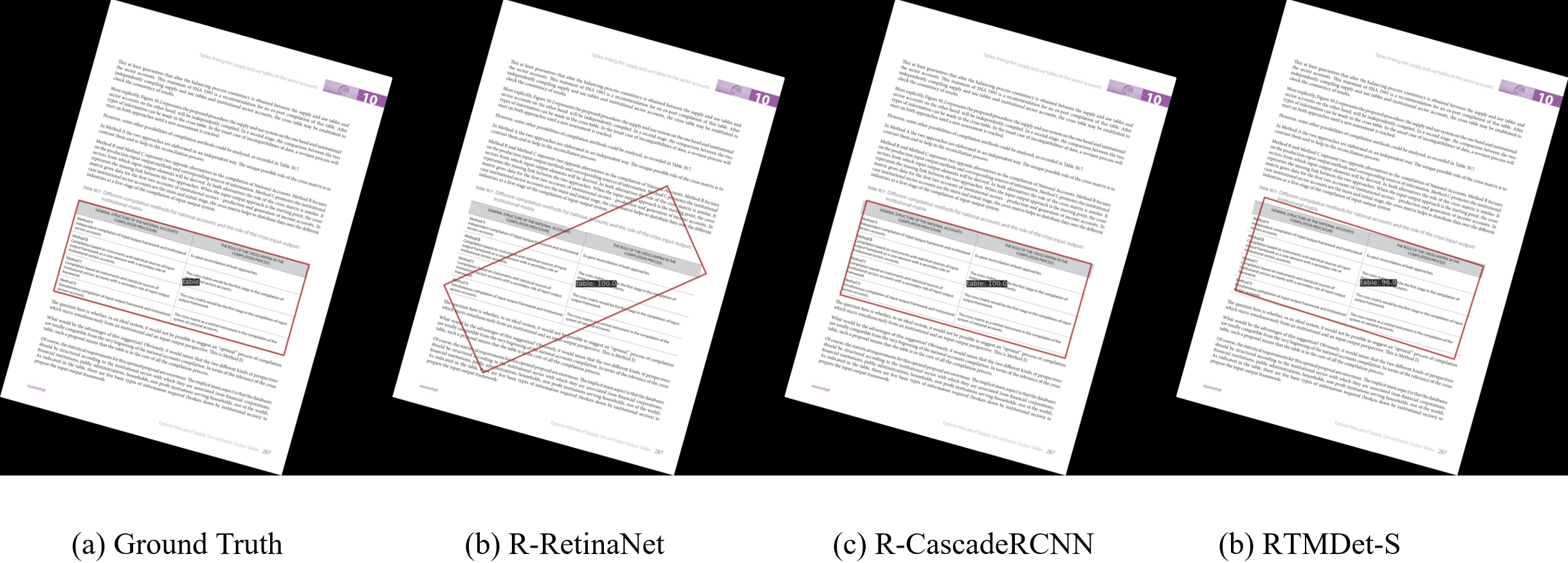}
\caption{Baseline Model Predict Visualization\label{fig:baselinevis}}
\end{figure}  

In this section, experiments were conducted on three models: R-RetinaNet, R-Cascade-RCNN, and RTMDet-S. The objective was to select an algorithm that strikes a balance between accuracy and speed, which will then serve as the baseline for implementing rotation table detection and head/tail recognition.

The results are displayed in Table \ref{tab:rotatebaseline}. R-RetinaNet achieved an AP50(T<360) score of 0.547, as depicted in the prediction outcomes in Figure \ref{fig:baselinevis}. However, the predictions were not precise and the speed was limited to only 13.9 FPS.

\begin{figure}[H]
\centering
\includegraphics[width=8.5 cm]{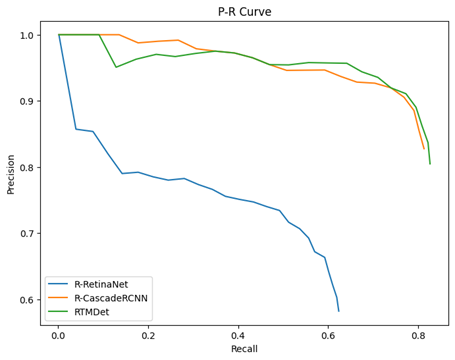}
\caption{Baseline P-R Curve on the TRR360D Dataset\label{fig:baselinepr}}
\end{figure}

R-CascadeRCNN achieved an AP50(T<360) score of 0.788. The PR curve illustrated in Figure \ref{fig:baselinepr} displayed superior performance compared to R-RetinaNet, and the visualization of its prediction outcomes in Figure \ref{fig:baselinevis} (c) demonstrated that the region predictions were predominantly accurate. Nonetheless, its speed was a mere 7.6 FPS, which was slower than R-RetinaNet. Although the accuracy was enhanced relative to R-RetinaNet, there was a trade-off with reduced speed.

As depicted in Figure \ref{fig:baselinevis} (d), Table \ref{tab:rotatebaseline}, and the PR curve in Figure \ref{fig:baselinepr}, RTMDet-S demonstrated a level of performance that is comparable to R-CascadeRCNN in terms of AP50(T<360) precision. Additionally, it boasted a speed of 33.7 FPS, which is significantly faster than R-RetinaNet and R-CascadeRCNN. Based on these results, the baseline model chosen for table rotation detection and head-tail recognition is RTMDet-S.

\begin{table}
\caption{Baseline model experimental results on the TRR360D Dataset\label{tab:rotatebaseline}}
\centering

\begin{tabular}{lll}
\toprule
\textbf{Model}	& \textbf{AP50(T<360)}	& \textbf{FPS}\\
\midrule
R-RetinaNet		& 0.547			& 13.9 \\
R-CascadeRCNN		& 0.788			& 7.6 \\
RTMDet-S		& 0.788			& 33.7 \\

\bottomrule
\end{tabular}

\end{table}

\subsection{Ablation Study on Head and Tail Location}

\begin{figure}[ht]
\centering
\includegraphics[width=1.0\textwidth]{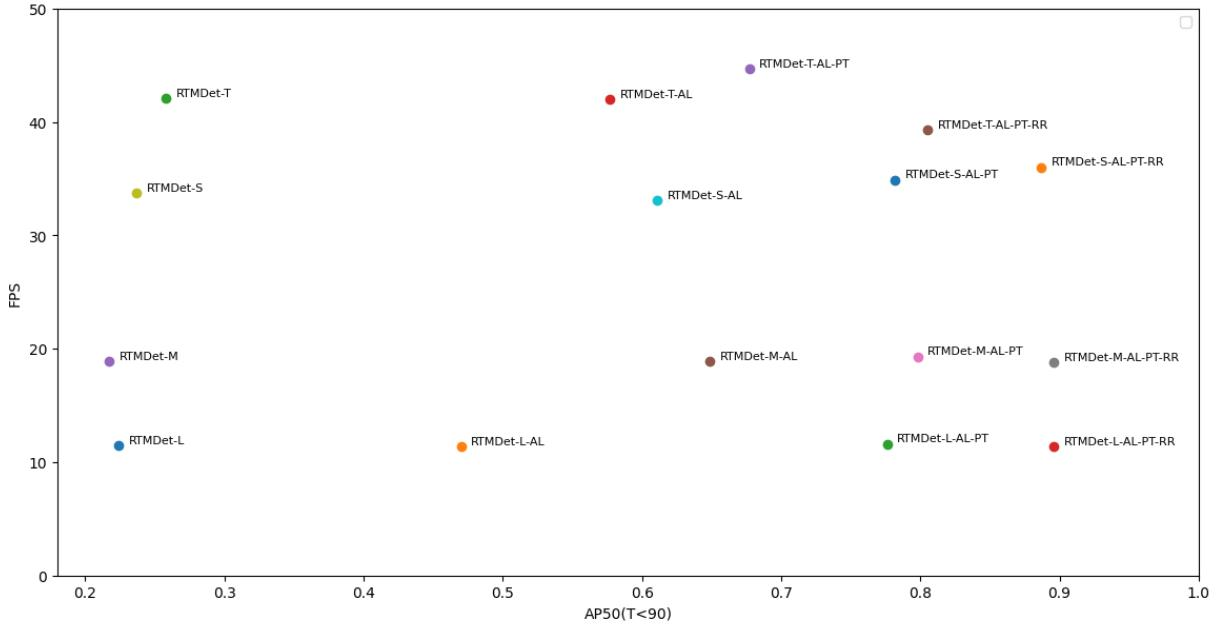}
\caption{FPS-AP50(T<90) on the TRR360D Dataset}
\label{fig:fpsap50}
\end{figure}

Based on the RTMDet baseline model, this section conducted extensive experiments on the TRR360D dataset, and the experimental results are shown in Table \ref{tab:rtmdeta}. In Table \ref{tab:rtmdeta}, L, M, S, and T in the Rotated RTMDet comparison experiment respectively represent the large, medium, small, and tiny models of RTMDet. Their FPS-AP chart is shown in Figure \ref{fig:fpsap50}, with the T series model having the fastest average FPS of 42.0, the S series model having a relatively fast average FPS of 34.4, the M series model having a moderate average FPS of 18.9, and the L series having the slowest average FPS of 11.5. In Table \ref{tab:rtmdeta} AL represents the Angle Loss branch, PT represents transfer learning, and RR represents adaptive boundary random rotation data augmentation.

The experimental reasoning test environment is based on the Ubuntu 20.04 system, and the hardware environment used is: 12th generation Intel i7-12700 processor, 32GB RAM, NVIDIA GeForce 3060 12G GPU. Different hardware platforms may have been used during the training phase, such as NVIDIA GeForce 3060 12G GPU, NVIDIA Tesla V100 GPU, etc. The training used the AdamW optimizer for 32400 iterations, with a batch size set to 2.

\subsubsection{AL Branch}

As shown in Table \ref{result} of the rotation RTMDet comparison experiment, the AP50(T<360) accuracy for RTMDet-L, RTMDet-M, RTMDet-S, and RTMDet-T is 0.787, 0.786, 0.788, and 0.778, respectively. When the absolute difference between the predicted bounding box angle and the ground truth angle is limited to less than 90°, the AP(T<90) accuracy is 0.224, 0.217, 0.237, and 0.258, representing a decrease of 56.3\%, 56.9\%, 55.1\%, and 52.0\%, respectively. This indicates that the original RTMDet in MMRotate can only predict the regions of rotated bounding boxes, but it cannot learn and predict the semantic features of the four corner points of the rotated bounding boxes. As a result, it fails to distinguish between the 30° and -150° rotated bounding boxes shown in Figure \ref{fig:r360}, and can only represent a range of 180°. This highlights the limitation of the original RTMDet-S and other related algorithms in not supporting the localization of the head and tail of rotated table regions.

Upon integrating the AL branch into the RTMDet model, the AP50(T<90) metrics of RTMDet-L-AL, RTMDet-M-AL, RTMDet-S-AL, and RTMDet-T-AL showed \textbf{an improvement of 24.6, 43.2, 37.4, and 31.9 percentage points}, respectively, compared to their respective counterparts RTMDet-L, RTMDet-M, RTMDet-S, and RTMDet-T. This result highlights the effectiveness of the AL branch in facilitating 360° rotation feature learning, enabling the detection of both the head and tail of tables. 

\begin{figure}[ht]
\centering
\includegraphics[width=16.5cm]{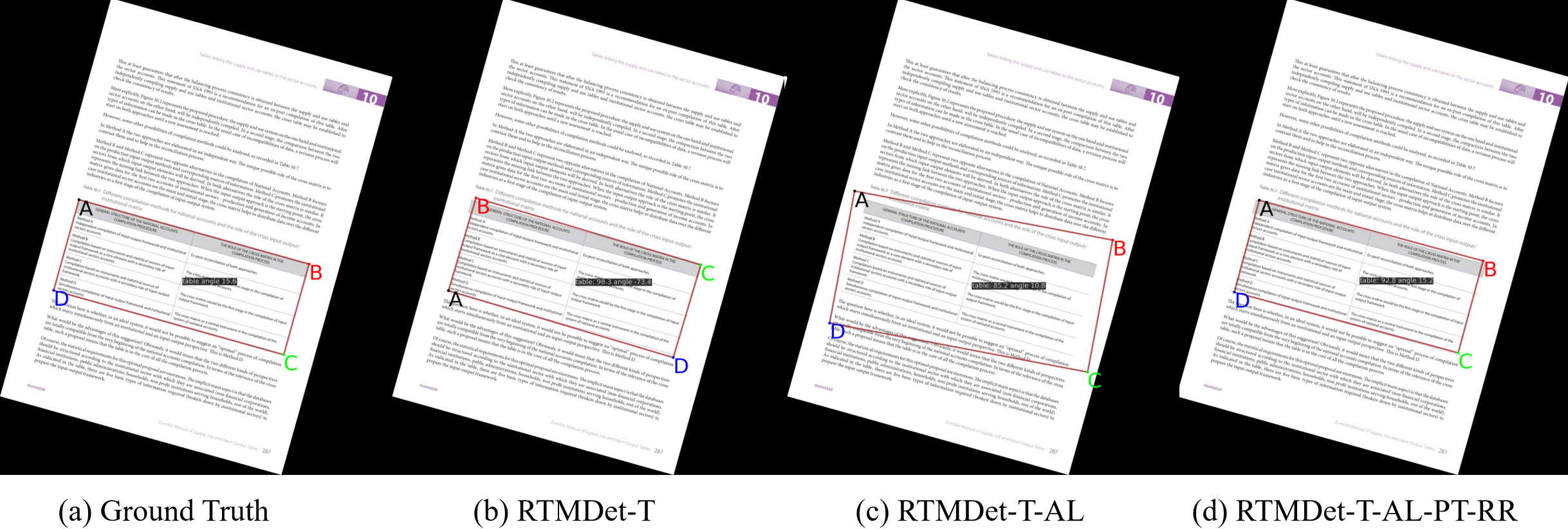}
\caption{RTMDet Predict Visualization\label{rthdetvis}}
\end{figure}

The visualization of the detection results also serves as evidence for the effectiveness of the AL branch. Figure \ref{rthdetvis} shows the visualization of the actual rotated bounding boxes for sample 10016 in the TRR360D rotated table detection test set. Specifically, Figure \ref{rthdetvis} (a) depicts the actual rotated bounding boxes, while Figure \ref{rthdetvis} (b) illustrates that although the RTMDet-L model correctly predicts the table region, the RIoU loss function fails to constrain the starting point of the table, leading to an inaccurate prediction of point A. By integrating the AL branch into the RTMDet-L model, the starting point prediction is effectively constrained, as shown in Figure \ref{rthdetvis} (c). While the prediction of point A is roughly correct, it has a certain impact on the accuracy of the predicted table region. Therefore, two methods are employed in the following to improve the accuracy.



\begin{table}
\caption{Comparison of RTMdet models with different configurations on the TRR360D Dataset\label{result}}
\centering
\begin{tabular}{llll}
\toprule
\textbf{Model} & \textbf{AP50 (T<360)} & \textbf{AP50 (T<90)} & \textbf{FPS}  \\
\midrule
RTMDet-L & 0.787 & 0.224 & 11.5 \\
RTMDet-L-AL & 0.566 & 0.470 & 11.4 \\
RTMDet-L-AL-PT & 0.793 & 0.776 & 11.6 \\
RTMDet-L-AL-PT-RR & 0.897 & 0.896 & 11.4\\
RTMDet-M & 0.786 & 0.217 & 18.9 \\
RTMDet-M-AL & 0.690 & 0.649 & 18.9 \\
RTMDet-M-AL-PT & 0.804 & 0.798 & 19.3 \\
RTMDet-M-AL-PT-RR & 0.896 & 0.896 & 18.8 \\
RTMDet-S & 0.788 & 0.237 & 33.7 \\
RTMDet-S-AL & 0.693 & 0.611 & 33.1 \\
RTMDet-S-AL-PT & 0.792 & 0.782 & 34.9 \\
RTMDet-S-AL-PT-RR & 0.890 & 0.887 & 36.0 \\
RTMDet-T & 0.778 & 0.258 & 42.1 \\
RTMDet-T-AL & 0.597 & 0.577 & 42.0 \\
RTMDet-T-AL-PT & 0.764 & 0.677 & 44.7 \\
RTMDet-T-AL-PT-RR & 0.805 & 0.805 & 39.3 \\
\bottomrule
\end{tabular}

\label{tab:rtmdeta}
\end{table}

\subsubsection{Transfer Learning}

In this study, Transfer Learning is also attempted to enhance the accuracy of RTHDet. Table \ref{result} presents the results of the rotating RTMDet comparison experiment. It shows that the RTMDet-L-AL-PT, RTMDet-M-AL-PT, RTMDet-S-AL-PT, and RTMDet-T-AL-PT models, which utilize transfer learning from the non-table detection domain DOTA dataset, outperform their counterparts without transfer learning, namely RTMDet-L-AL, RTMDet-M-AL, RTMDet-S-AL, and RTMDet-T-AL, in terms of AP50(T<90) metrics. Specifically, the AP50(T<90) index for the former models increased by 30.6, 14.9, 17.1, and 10.0 percentage points, respectively. These results demonstrate the effectiveness of transfer learning in the rotating table detection domain.

\subsubsection{Adaptive Boundary Random Rotation Data Augmentation}

The Adaptive Boundary Rotation method \ref{subsection:AdaptivelyBoundedRotation} is applied to augment the training data on the unrotated horizontal table training set of TRR360D. This approach enriches the data and improves the model's accuracy. Table \ref{result} of the rotating RTMDet comparison experiment shows that the RTMDet-L-AL-PT-RR, RTMDet-M-AL-PT-RR, RTMDet-S-AL-PT-RR, and RTMDet-T-AL-PT-RR models, which use random rotation data augmentation with adaptive boundaries, outperformed their counterparts without this augmentation technique (i.e., RTMDet-L-AL-PT, RTMDet-S-AL-PT, RTMDet-S-AL-PT, and RTMDet-T-AL-PT) in terms of the AP50(T<90) metric, with an increase of 12.0, 9.8, 10.5, and 12.8 percentage points, respectively. The inference visualization of the RTMDet-L-AL-PT-RR model shown in Figure \ref{rthdetvis} (d) meets the accuracy requirements of region localization and correctly predicts the starting point A at the top-left corner of the table. As a result, line segment AB represents the head of the table, while line segment DC represents the tail of the table. These results demonstrate that adaptive boundary random rotation data augmentation effectively improves the performance of the RTHDet model.

\section{Conclusion}

Traditional models and datasets primarily concentrated on horizontal table detection, unable to accurately detect table regions and localize their head and tail parts in rotating scenarios, thus greatly restricting the development of table recognition in rotated contexts. Therefore, this paper proposed a novel task and provided substantial innovations in detecting table regions and localizing their head and tail parts in rotational scenarios, along with the proposal of datasets, evaluation metrics, and methods. The Adaptively Bounded Rotation method is proposed to generate the TRR360D dataset on the classic ICDAR2019MTD dataset. The newly introduced R360 AP evaluation metric significantly improved the assessment of the precision of rotated region detection. Leveraging the $D_{r360}$ angle definition method, Transfer Learning, and Adaptive Boundary Rotation data augmentation on the RTMDet-S baseline model, the proposed RTHDet remarkably improved AP50 (T<90) from 23.7\% to 88.7\%, thereby attesting to its efficacy in addressing this innovative task. This research is poised to propel the development of downstream table recognition OCR tasks. Despite the significant progress, the current inability to detect arbitrary quadrilateral tables in perspective scenarios remains a notable limitation, necessitating further research to address this challenge.

\bibliographystyle{unsrt}  
\bibliography{references}

\end{document}